\documentclass[letterpaper, 10 pt, conference]{ieeeconf}  

\usepackage{booktabs}
\usepackage{multirow}
\usepackage{amsmath}
\usepackage{amssymb}
\usepackage{graphicx}
\usepackage{cite}
\usepackage{booktabs}
\usepackage{multirow}
\usepackage{capt-of}
\IEEEoverridecommandlockouts                              

\usepackage{graphics} 

\title{\LARGE \bf
VisForce: Visual Grounding of Current and Desired Forces for Goal-Conditioned Dexterous Manipulation
}

\author{Jung-Woo Lee, Soo-Chul Lim$^{*}$}

\begin{document}

\maketitle
\thispagestyle{empty}
\pagestyle{empty}

\begin{abstract}
Vision-Language-Action (VLA) models have emerged as general-purpose robotic manipulation policies. However, in dexterous hand manipulation, contact forces are typically provided as separate states or force-specific representations, making it difficult to explicitly represent the spatial correspondence between force and their corresponding visual locations. In this work, we propose VisForce, which visually grounds the current and desired forces at their corresponding fingertip locations. VisForce renders current and desired visual force cues on the current wrist image and a task-specific goal image, and combines the two representations through goal-conditioned cross-attention to generate force-aware actions. We evaluate VisForce using a real UR10 robot equipped with an RH56F1 dexterous hand through force-conditioned grasping and three multi-stage manipulation tasks. In force-conditioned grasping experiments, VisForce exhibited a consistent grip-force response as the desired force increased, and achieved grasp-and-lift success rates of 70\% and 80\% for an egg and a toothpaste tube, respectively. It further achieved final success rates of 70\%, 55\%, and 40\% on cup insertion/bottle pouring, tong-assisted bread transfer, and slip-modulated peg-in-hole, respectively. These results show that fingertip-aligned visual force representations can be effectively used for force-aware conditioning in VLA-based dexterous hand manipulation.
\end{abstract}

\begin{figure*}[!t]
    \centering
    \centerline{\includegraphics[width=\textwidth]{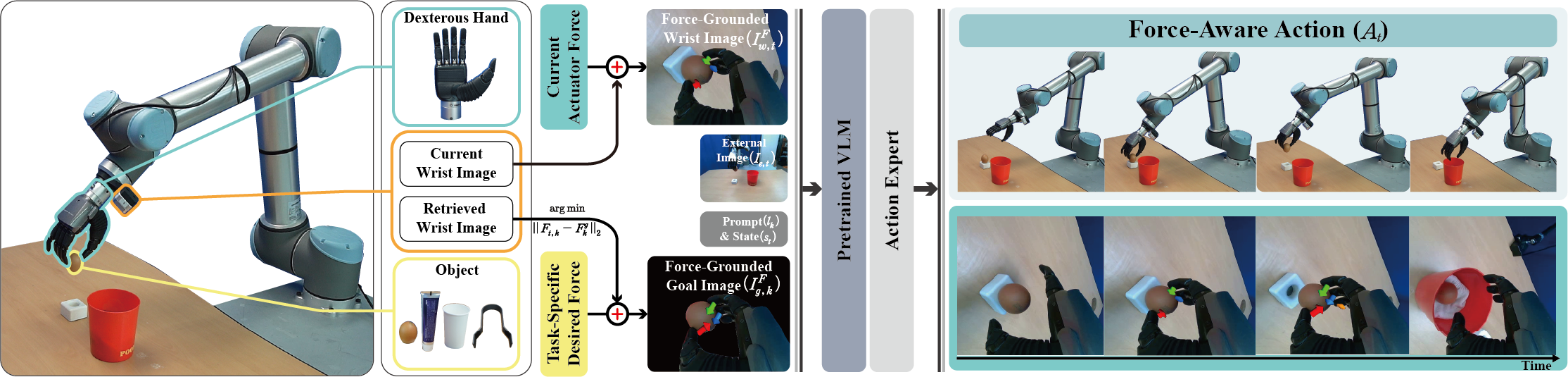}}
    \caption{\textbf{Overview of the proposed VisForce framework.} VisForce represents the actuator force and task-specific desired force in a common visual space by grounding both signals at their corresponding fingertip locations. The actuator force is overlaid on the current wrist image, while the desired force is rendered on the retrieved wrist image. These force-grounded current and goal observations, together with the external image, robot state, and language instruction, are used to condition the pretrained VLM and action expert for force-aware action generation.}
    \label{fig:overview} 
\end{figure*}

\section{INTRODUCTION}

Vision-Language-Action (VLA) models have emerged as general-purpose policies for generating diverse robotic manipulation behaviors from visual observations and natural-language instructions by leveraging large-scale vision-language pretraining~\cite{pi_0, pi_05, octo, rt-2, openvla}.

However, in contact-rich manipulation with dexterous hands, both the hand configuration and the applied contact force are important. For example, when grasping deformable objects, insufficient or excessive force can lead to slippage or object deformation, respectively. Such force-dependent interactions are difficult to explicitly condition on using only the visual and proprioceptive observations commonly employed in existing VLA models.

To address this limitation, recent studies have incorporated force or tactile signals into VLA models~\cite{forcevla, taf_vla, fd_vla, tactile_vla, vla_touch, forcesight}, while others have used goal images or visual subgoals to specify desired manipulation states more concretely~\cite{cot, goalrep, susie, forediffusion, foreact, goal_vla}. However, force-aware approaches typically process force through separate state vectors, latent tokens, or dedicated fusion pathways, while representations in which force is expressed as a visual condition spatially aligned with the corresponding fingertip remain limited. On the other hand, visual-goal approaches provide desired hand/object configurations or future scenes, but generally do not include desired force as an explicit conditioning signal. As a result, two aspects that are both important in dexterous grasping—``how to grasp'' and ``how hard to grasp''—are often encoded in separate representations.

In human grasping, anticipatory hand preshaping occurs prior to contact based on the visual properties of an object~\cite{schettino2003effects}, while grip force is also predictively regulated based on prior experience and sensory cues~\cite{johansson2009coding}. These findings suggest the potential importance of jointly considering hand configuration and force regulation when planning interactions with objects.

In this work, we propose VisForce, which provides hand configuration and force conditions within a unified visual representation. Fig.~\ref{fig:overview} presents an overview of the proposed framework. VisForce spatially grounds the current actuator forces of individual fingers at their corresponding fingertip locations in the wrist image, allowing force information to be processed through the existing visual pathway of a VLA model without a separate force-specific encoder or latent force token. In parallel, task-specific desired forces are grounded at the corresponding fingertip locations in a foreground-isolated goal image, jointly representing the desired hand/object configuration and force condition. The force-grounded current and goal representations are then connected through goal-conditioned cross-attention, where current wrist tokens attend to task-relevant foreground goal tokens. The resulting representation, together with the external-camera image, language instruction, and robot state, is provided to the pretrained VLM and action expert for force-aware action generation.

The main contributions of this work are as follows:
\begin{itemize}
    \item We introduce a force-grounded current visual representation that spatially aligns per-finger actuator forces with their corresponding fingertip locations.
    \item We propose a force-grounded goal representation that unifies task-relevant hand/object configuration and finger-wise desired forces.
    \item We incorporate goal-conditioned cross-attention to couple the current force-grounded observation with the desired goal representation for force-aware action generation.
\end{itemize}

\section{RELATED WORK}

\subsection{Vision-Language-Action Models}

VLA models leverage pretrained vision-language representations to generate robot actions from visual observations and language instructions~\cite{rt-2, openvla, pi_05}. Recent work has extended this paradigm to dexterous and multi-arm manipulation, including DexGraspVLA~\cite{dexgraspvla} and PEAfowl~\cite{fan2026peafowl}. As VLA models expand toward contact-rich manipulation, representing physical interaction states becomes increasingly important. VisForce addresses this gap by reformulating actuator force as an image-space representation that can be processed through the visual pathway of a VLA model.

\subsection{Force- and Tactile-aware Manipulation}

Force and tactile feedback have been incorporated into dexterous and contact-rich manipulation through tactile sensing, multimodal sensor fusion, force-informed policy learning, and reactive control~\cite{dextouch, dexforce, kinedex, visual_force_tactile, foar, implicitrdp, pocodp3}. VLA-based approaches further integrate force or tactile information through force-aware experts, adapters, latent force representations, or tactile-conditioned controllers~\cite{forcevla, taf_vla, fd_vla, vla_touch}. These approaches typically encode force or tactile signals as separate modalities or learned representations that are fused with visual features. ForceSight~\cite{forcesight} more explicitly associates visual observations with target fingertip locations and force goals. In contrast, VisForce grounds actuator-level force measurements onto their corresponding fingertip locations in the visual observation, explicitly representing the spatial relationship between force and hand configuration in the input image.

\subsection{Goal-conditioned Policy and Visual Foresight}

Goal-conditioned manipulation methods use goal images, learned visual goal
representations, or generated visual subgoals to provide task specifications
beyond language alone~\cite{goalrep, susie, grmg}. Related visual planning and foresight approaches generate goal or subgoal images, or predict future observations, to provide intermediate guidance for manipulation~\cite{cot, foreact, forediffusion, goal_vla}. VisForce extends visual goal conditioning by spatially grounding task-specific desired forces at the corresponding fingertip locations in the goal image, allowing the policy to reason jointly about the desired visual configuration and force state.

\section{METHODS}

\begin{figure*}[t]
    \centering
    \centerline{\includegraphics[width=\textwidth]{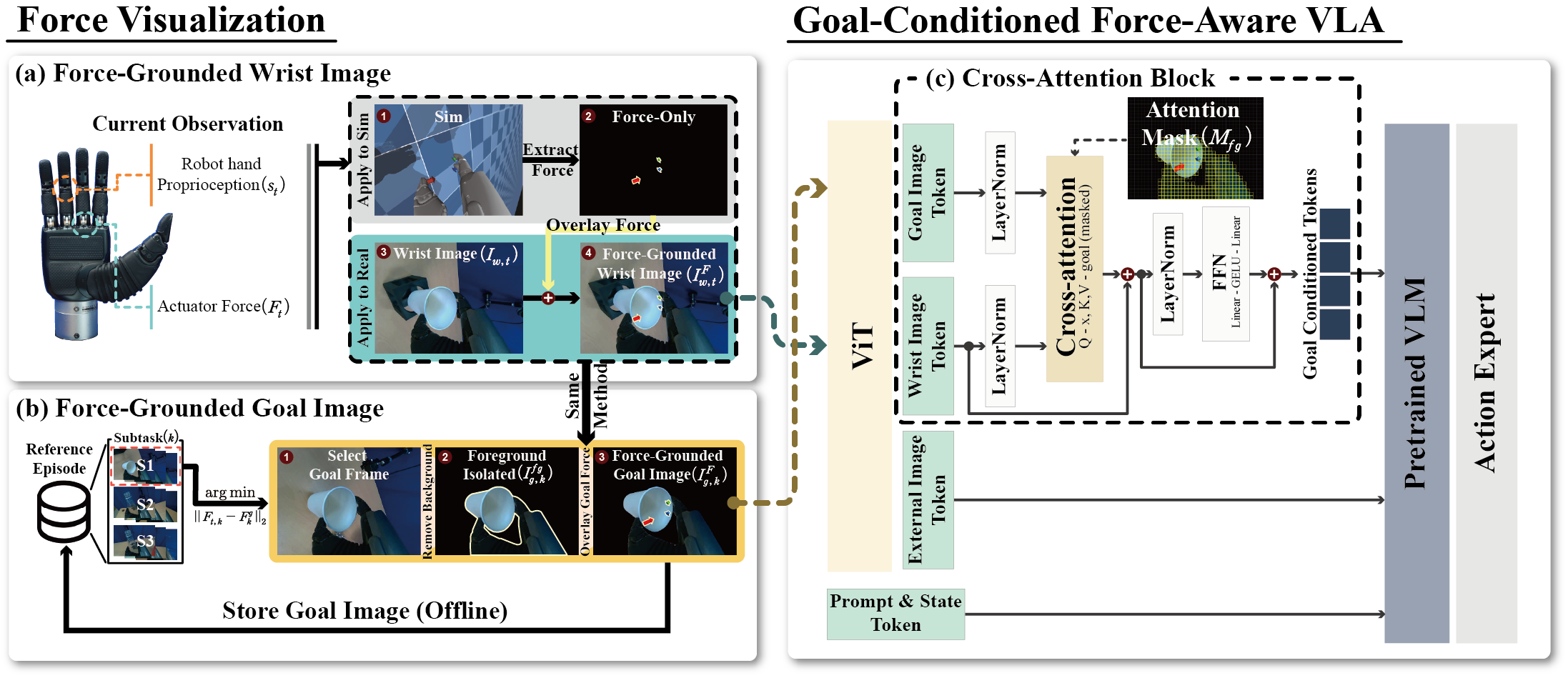}}
    \caption{\textbf{VisForce Architecture: Force-Grounded Visual Construction and Goal-Conditioned VLA.} Left: fingertip-aligned visual grounding of current and desired actuator forces for wrist and goal image construction. Right: masked goal-conditioned cross-attention integrates current wrist and foreground goal representations for force-aware action generation.}
    \label{fig:method} 
\end{figure*}

Our method builds on $\pi_{0.5}$~\cite{pi_05}, and introduces two components, as illustrated in Fig.~\ref{fig:method}: (a,b) force-grounded visual representations of the current observation and goal, and (c) goal-conditioned cross-attention for their fusion.

\subsection{Problem Formulation}

We consider the problem of force-aware manipulation with a dexterous hand. At timestep $t$, the policy observes the proprioceptive state $\mathbf{s}_t$, wrist and external images $I_{w,t}$ and $I_{e,t}$, subtask instruction $\ell_k$, and finger-wise actuator forces $\mathbf{F}_t=[f_t^1,\ldots,f_t^5]\in\mathbb{R}^5$. The current subtask $k$ is specified by the user.

The wrist image $I_{w,t}$ and finger-wise actuator forces $F_t$ are combined to construct the force-grounded wrist image $I^F_{w,t}$. We additionally use the force-grounded goal image $I^F_{g,k}$ corresponding to the current subtask $k$ as a visual condition. As shown in Fig.~\ref{fig:overview}, the policy takes $I^F_{w,t}$, $I_{e,t}$, $I^F_{g,k}$, $s_t$, and $\ell_k$ as input and generates an $H$-step action chunk $A_t=[a_t,a_{t+1},\ldots,a_{t+H-1}]\in\mathbb R^{H\times12}$:

\begin{equation}
A_t \sim
\pi\left(
\cdot \mid
I^F_{w,t},
I_{e,t},
I^F_{g,k},
s_t,
\ell_k
\right).
\label{eq:policy_define}
\end{equation}

Each action $a_t$ is defined as a 12-dimensional delta action,
$a_t=[\Delta a^{\mathrm{arm}}_t, \Delta a^{\mathrm{hand}}_t]$, where
$\Delta a^{\mathrm{arm}}_t \in \mathbb{R}^{6}$ denotes the delta joint command for the six joints of the robot arm, and
$\Delta a^{\mathrm{hand}}_t \in \mathbb{R}^{6}$ denotes the delta joint command for the six joints of the dexterous hand.

\subsection{Force Visualization via Wrist-Camera Aligned Rendering}

Each actuator force $f_t^i$ is a scalar measurement without explicit spatial correspondence to its associated fingertip, which can lead to ambiguity under fingertip occlusion. To address this, we reproduce the real robot configuration in simulation and render each actuator force as a visual cue at the corresponding fingertip location. After aligning the simulated wrist camera with the real wrist camera, the rendered force cues are overlaid onto the real wrist image to construct the force-grounded wrist image.

\textbf{Wrist-camera alignment.}
The force cues are rendered as 3D geometry in MuJoCo~\cite{todorov2012mujoco} and then overlaid onto the real wrist image. To align the rendered cues with the corresponding real fingertip locations, we first align the 6-DoF extrinsic pose of the simulated wrist camera with the real wrist camera before force rendering.

The simulated wrist camera is aligned to the real camera by optimizing its 6-DoF extrinsic pose using a combination of hand-mask IoU~\cite{rahman2016iou} and edge-based Chamfer losses~\cite{borgefors1988chamfer}. Real masks are obtained using GroundingDINO~\cite{groundingdino} and SAM2~\cite{sam2}, while simulated masks are rendered in MuJoCo. We use a coarse-to-fine optimization strategy with Differential Evolution~\cite{storn1997differential} followed by Powell optimization~\cite{powell1964efficient}.

Once camera alignment is completed, we reproduce the real robot joint state in MuJoCo at each timestep $t$ and compute the 3D position $p_t^i$ and normal direction $d_t^i$ of the $i$-th fingertip using hand kinematics. To spatially ground the force magnitude at the corresponding fingertip location, we use $d_t^i$ as the rendering direction and define

\begin{equation}
v_t^i = \alpha f_t^i d_t^i
\label{eq:force_render}
\end{equation}

where $\alpha$ scales the force magnitude to the rendered vector length; we use $\alpha=1\times10^{-4}$. Thus, $v_t^i$ is not an estimated contact-force vector, but a visual force cue that associates the actuator-force magnitude with the corresponding fingertip location.

\textbf{Force rendering and overlay.}
As illustrated in Fig.~\ref{fig:method}(a), in step (1), we create a 3D arrow geometry from $p_t^i$ to $p_t^i+v_t^i$ using a finger-specific color in the aligned simulated wrist-camera view. In step (2), we render a force-only image containing only the visual force cues generated in (1). Finally, in steps (3)--(4), we overlay the force-only image onto the real wrist image $I_{w,t}$ to construct the force-grounded wrist image $I^F_{w,t}$. To avoid unnecessary cues caused by small actuator fluctuations, visual force cues are rendered only for forces above a predefined threshold.

\subsection{Force-Grounded Goal Image Construction}

For each subtask $k$, we construct a force-grounded goal image $I^F_{g,k}$ that jointly represents the desired hand/object configuration and fingertip-aligned desired forces. Since conventional visual goals do not explicitly encode the desired force at each fingertip, we ground the desired forces at their corresponding fingertip locations and isolate the task-relevant foreground.

As illustrated in Fig.~\ref{fig:method}(b), the goal image is constructed in three steps. (1) We select one demonstration from each task as a reference episode, and choose as the representative goal frame the frame within the subtask $k$ segment whose measured actuator force is closest to the predefined desired force $F_k^g$. (2) From the corresponding wrist image, we segment the robot hand and manipulated object using GroundingDINO and SAM2, and remove the background to obtain a foreground-isolated image $I^{\mathrm{fg}}_{g,k}$. (3) We then reproduce the hand joint configuration of the selected frame in MuJoCo and render the desired force $F_k^g$ at the corresponding fingertip locations using the same rendering procedure as in the current force visualization, thereby constructing the final force-grounded goal image $I^F_{g,k}$.

The resulting goal images $I^F_{g,k}$ are stored offline and retrieved by subtask during inference; segmentation is therefore not required online.

\subsection{Goal-Conditioned Force-Aware VLA}

To jointly condition the policy on the force-grounded wrist image $I^F_{w,t}$ and the force-grounded goal image $I^F_{g,k}$, we introduce a single cross-attention block. Before applying cross-attention, the force-grounded wrist image $I^F_{w,t}$, force-grounded goal image $I^F_{g,k}$, and external-camera image are encoded by a shared SigLIP visual encoder into visual tokens in the same representation space.

We then apply goal-conditioned cross-attention so that the current wrist-image tokens attend to task-relevant goal tokens. To exclude background tokens from the key and value inputs of cross-attention, we compute a patch-level foreground mask $M_{\mathrm{fg}}$. As illustrated by the Attention Mask in Fig.~\ref{fig:method}(c), a $14\times14$ patch is defined as a foreground patch if it contains at least one pixel whose value differs from the background value. The mask $M_{\mathrm{fg}}$ restricts the key and value inputs of cross-attention to only those goal tokens selected in this way.

As illustrated in Fig.~\ref{fig:method}(c), the GoalAttnBlock consists of a pre-LayerNorm cross-attention block with 8 attention heads. The current force-grounded wrist-image tokens $x_t$ are used as queries, while the force-grounded goal-image tokens $c_k$ serve as keys and values to compute the goal-conditioned update. The wrist-image representation is then updated through a residual connection followed by a two-layer FFN with an expansion ratio of 4 and GELU activation:

\begin{figure*}[t]
    \centering
    \centerline{\includegraphics[width=\textwidth]{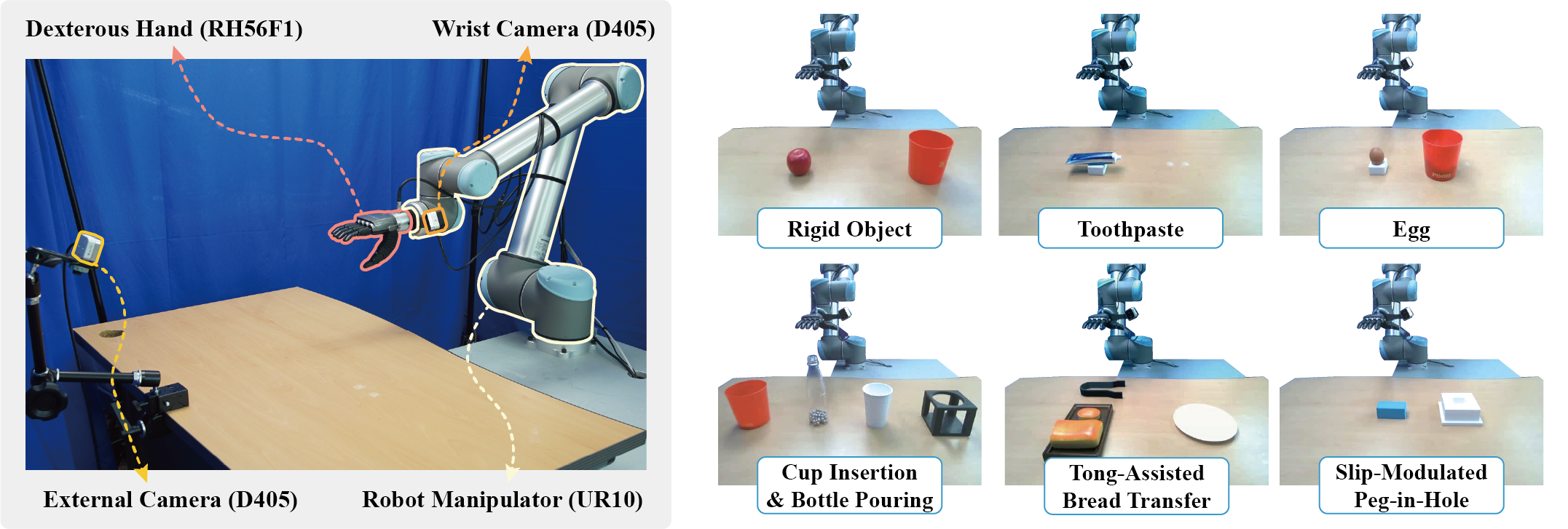}}
    \caption{\textbf{Real-World Experimental Setup and Evaluation Tasks.} Left: real-world manipulation platform composed of a UR10 robot arm, an RH56F1 dexterous hand, a wrist-mounted D405 camera, and an external D405 camera. Right: task set used for evaluation. The top row shows the force-conditioned grasping tasks used in T1 (rigid object, toothpaste, and egg), and the bottom row shows the multi-stage force-aware manipulation tasks used in T2--T4 (cup insertion and bottle pouring, tong-assisted bread transfer, and slip-modulated peg-in-hole).}
    \label{fig:experiment_setup} 
\end{figure*}

\begin{equation}
\begin{aligned}
{x^{\prime}}_t
&=
x_t +
\mathrm{CA}\!\left(\mathrm{LN}(x_t), \mathrm{LN}(c_k); M_{\mathrm{fg}}\right), \\
x^G_t 
&= 
{x^{\prime}}_t +
\mathrm{FFN}\!\left(\mathrm{LN}({x^{\prime}}_t)\right).
\end{aligned}
\label{eq:goal_attn}
\end{equation}

where $x_t^G$ denotes the goal-conditioned current visual representation. Goal-conditioned cross-attention is applied only to the wrist-image stream, while the external-camera tokens are processed by the shared SigLIP encoder and then follow the original visual pathway of $\pi_{0.5}$.

The updated representation $x_t^G$ is passed to the VLM backbone and action expert together with the external-camera image tokens, robot proprioception, and language instruction to generate the action chunk $A_t$. The training objective uses the same flow-matching loss as the base model.

During training, all parameters of the shared SigLIP visual encoder and the goal-conditioned cross-attention module are updated, while the pretrained VLM backbone and action expert are fine-tuned using LoRA~\cite{hu2021lora}. The same LoRA configuration is applied to all comparison models, with rank $r=16$ and scaling factor $\alpha=16$.

\section{EXPERIMENTS AND RESULTS}

We evaluate VisForce from four perspectives to examine how effectively it supports diverse force-dependent interactions required in real-world dexterous hand manipulation. Specifically, we evaluate (1) whether changes in the desired force condition lead to corresponding modulation of the actual actuator force, (2) whether spatially grounding force at the corresponding fingertip locations helps distinguish grasp states even when the fingertips are occluded by the object, (3) whether the policy can repeatedly adjust the required force level across manipulation stages while maintaining a persistent tool grasp, and (4) whether it can induce controlled slip through force modulation and subsequently transition to a stable grasp. These four evaluation aspects correspond to T1–T4, respectively.

\subsection{Experimental Setup}

\textbf{Hardware Setup.}
As shown in Fig.~\ref{fig:experiment_setup}, all experiments were conducted on a real robotic platform. The platform consists of a Universal Robots UR10 6-DoF robotic arm and an Inspire RH56F1 6-DoF dexterous hand. The camera system consists of a wrist-mounted camera and an external camera, both of which are Intel RealSense D405 models. Robot control and camera input are handled by the experimental control PC, while policy inference is performed on a separate workstation equipped with an NVIDIA RTX A6000 GPU.

\textbf{Data Collection.}
For each demonstration, we recorded wrist-camera images, external-camera images, proprioceptive states, actions, actuator forces, and task-specific desired forces. Camera images were recorded at 30 Hz, whereas robot states, actions, and forces were recorded at 200 Hz. The two streams were synchronized to the camera timestamps using nearest-neighbor matching. We collected 30, 25, 30, and 30 demonstrations for T1--T4, respectively.

\textbf{Comparison Methods.}
All comparison conditions use the same training demonstrations, action representation, and backbone. The main differences between the conditions lie in how the current and desired forces are represented and whether goal-conditioned cross-attention is used. This setup allows us to compare the effects of different force-conditioning strategies under the same training data and backbone.

\begin{itemize}
    \item $\pi_{0.5}$: The base policy without force information or an additional visual goal.
    
    \item State Force + Text Force Goal: The actuator force is appended to the robot state. The desired hand/object configuration is provided as a foreground-isolated goal image without visual force cues, while the desired force is specified numerically in the language prompt.
    
    \item Visual Force + Text Force Goal: The actuator force is provided through the force-grounded wrist image. The goal image and desired force are provided in the same manner as in State Force + Text Force Goal.
    
    \item VisForce w/o CA: The actuator force is provided through the force-grounded wrist image, and the desired force is rendered in the force-grounded goal image. However, the proposed cross-attention is not used; instead, the encoded wrist and goal visual tokens are directly concatenated and passed to the VLM backbone.
    
    \item VisForce (Ours): The same force-grounded wrist and goal images are used, while the proposed goal-conditioned cross-attention is applied so that the current wrist tokens can explicitly attend to the foreground goal tokens.
\end{itemize}

\textbf{Inference Protocol.}
During inference, the user specifies only the current subtask $k$. The corresponding language instruction and force-grounded goal image $I^F_{g,k}$ are then automatically retrieved and provided to the policy. Since T1 is a single-stage task, it uses a fixed task condition without any subtask transition, whereas in T2--T4, the subtask corresponding to each manipulation stage is specified.

\textbf{Evaluation Metrics.}
Each policy is evaluated over $N=20$ independent trials per task. For T2--T4, stage-wise success is defined as the cumulative success rate requiring all preceding stages to be completed successfully.

\begin{table}[!t]
\caption{Finger-wise desired force conditions [N] for T1 force-conditioned grasping.}
\label{tab:finger_desired_force_t1}
\resizebox{\columnwidth}{!}{
\begin{tabular}{llcccc}
\toprule
Task & Condition & Thumb & Index & Middle & Ring \\
\toprule
\multirow{3}{*}{T1: Rigid Object}
& Low force grasp & 0.7 & 0.5 & 0.5 & - \\

& Mid force grasp & 1.4 & 1.0 & 1.0 & - \\

& High force grasp & 2.0 & 1.5 & 1.5 & - \\

\midrule
\multirow{3}{*}{T1: Egg}
& Low force grasp & 0.12 & 0.12 & 0.12 & - \\

& Mid force grasp & 0.4 & 0.4 & 0.4 & - \\

& High force grasp & 1.0 & 1.0 & 1.0 & - \\

\midrule
\multirow{3}{*}{T1: Toothpaste}
& Low force grasp & 0.15 & 0.15 & 0.15 & - \\

& Mid force grasp & 0.4 & 0.4 & 0.4 & - \\

& High force grasp & 0.8 & 0.8 & 0.8 & - \\
\bottomrule
\end{tabular}}
\end{table}

\begin{table}[!t]
\caption{Finger-wise desired force conditions [N] for multi-stage manipulation tasks.}
\label{tab:finger_desired_force_multi}
\resizebox{\columnwidth}{!}{
\begin{tabular}{llcccc}
\toprule
Task & Subtask & Thumb & Index & Middle & Ring \\
\toprule
\multirow{3}{*}{T2}
& Cup grasping and insertion & 0.4 & 0.2 & 0.2 & - \\

& Bottle grasping and pouring & 1.2 & 0.8 & 0.8 & 0.8 \\

& Empty bottle placement & 0.6 & 0.4 & 0.4 & 0.4 \\
\midrule
\multirow{3}{*}{T3}
& Tong grasping / Bread placement & 0.4 & 0.4 & 0.4 & 0.2 \\

& Large bread transfer & 1.0 & 0.9 & 0.7 & 0.2 \\

& Small bread transfer & 1.5 & 1.2 & 1.0 & 0.2 \\
\midrule
\multirow{2}{*}{T4}
& Peg grasping and uprighting & 0.13 & 0.13 & 0.13 & - \\

& Peg insertion & 0.4 & 0.4 & 0.4 & - \\
\bottomrule
\end{tabular}}
\end{table}

\subsection{Evaluation Tasks}

\textit{T1: Force-Conditioned Grasping.}
T1 evaluates whether changes in the desired force are reflected in the policy's grasp behavior. In the Rigid Object setting shown in Fig.~\ref{fig:experiment_setup}, the object and pick-and-place task configuration are kept fixed while the desired force is varied across three levels: low, medium, and high. The hand--object configuration remains visually similar across conditions, with the rendered desired-force cue being the primary varying factor. Because the rigid object exhibits minimal visual deformation under changes in grip force, this setting allows us to evaluate actuator-force modulation with respect to the desired force while reducing the effects of object deformation and task variation.

We further evaluate whether this force modulation is effective for objects that require appropriate grasp forces, such as an egg and a toothpaste tube. In the Egg task, the robot lifts the object and places it into a red container, while in the Toothpaste task, it grasps and lifts the tube. The low-force condition in Table~\ref{tab:finger_desired_force_t1} is set to induce slip, whereas the high-force condition is set to cause egg breakage or leakage of the toothpaste contents. The training data include low-, medium-, and high-force demonstrations in equal proportions, with the medium condition used as the task-appropriate force for each object.

\textit{T2: Cup Insertion and Bottle Pouring.}
T2 corresponds to the Cup Insertion \& Bottle Pouring task shown in Fig.~\ref{fig:experiment_setup}. The robot performs cup insertion, bottle pouring, and bottle placement using the stage-dependent desired forces specified in Table~\ref{tab:finger_desired_force_multi}. During the bottle power grasp, most fingertips are occluded in the wrist-camera view. Therefore, T2 evaluates whether fingertip-aligned force grounding helps reduce ambiguity in the grasp state under such occlusion and supports stable execution of the subsequent manipulation.

\begin{figure*}[t]
    \centering
    \centerline{\includegraphics[width=\textwidth]{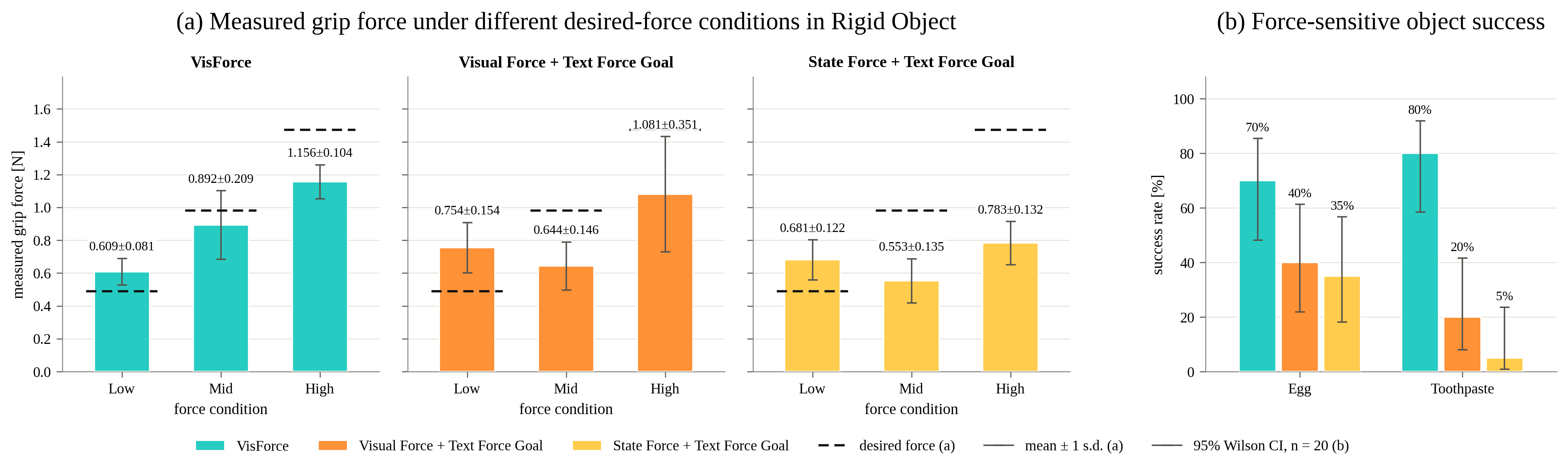}}
    \caption{\textbf{Force-Conditioned Grasping evaluation.}
    (a) Measured aggregate actuator force under low, medium, and high desired-force conditions for the rigid object. Bars indicate the mean over trials, error bars denote standard deviation, and dashed lines indicate the corresponding desired-force levels.
    (b) Grasp-and-lift success rates for the egg and toothpaste. Error bars denote 95\% Wilson confidence intervals. A trial is considered successful when the object is lifted without slip or dropping and without excessive-force-induced damage or deformation.}
    \label{fig:force_grasping} 
\end{figure*}

\begin{figure*}[!t]
\centering
\begin{minipage}[t]{0.49\textwidth}
    \vspace{0pt}
    \centering

    \includegraphics[width=\linewidth]{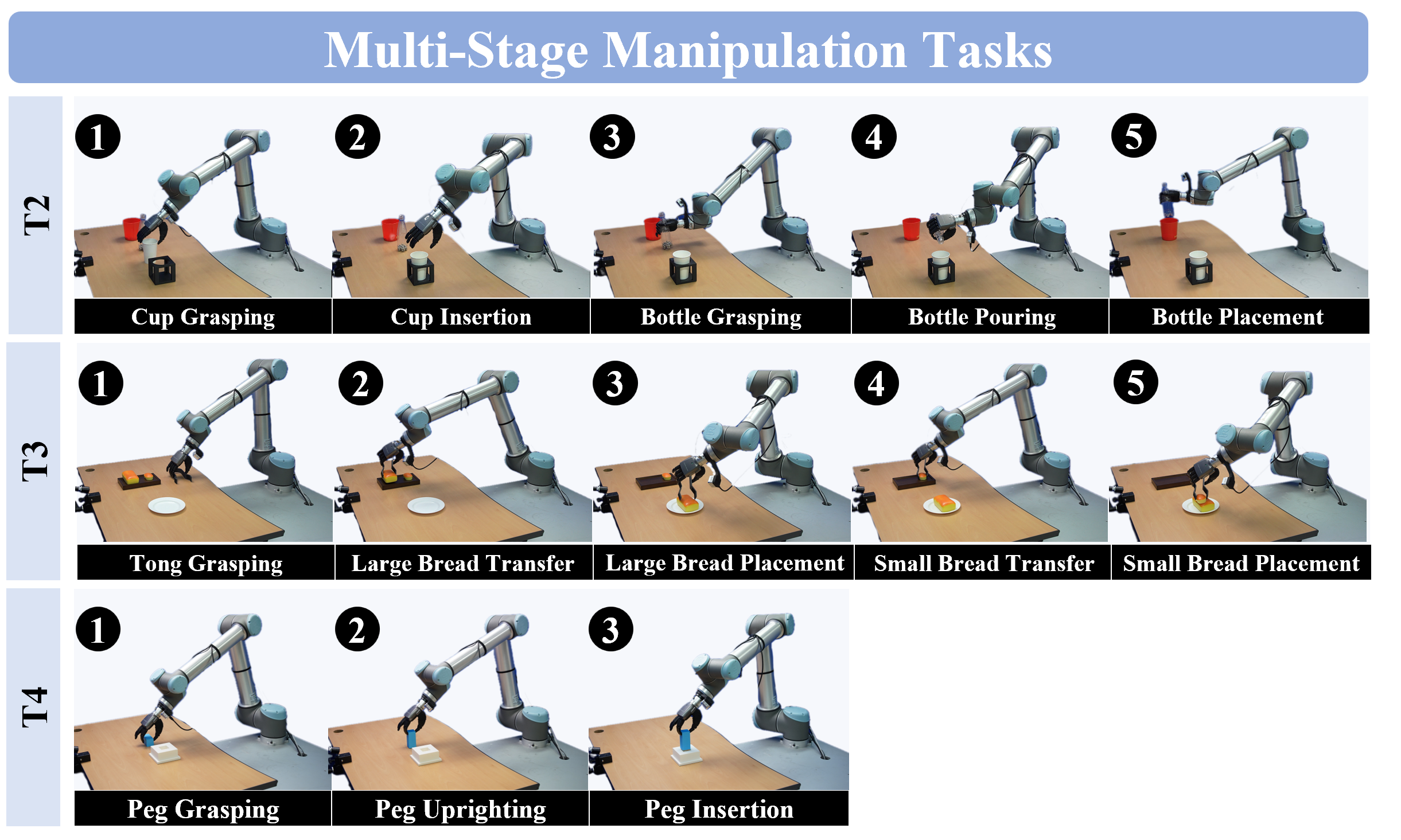}

    \captionof{figure}{
    \textbf{Stage-wise Evaluation of Multi-Stage Force-Aware Manipulation Tasks.} Sequential stages of T2--T4 used for cumulative success evaluation.
    }
    \label{fig:subtask_progress}
\end{minipage}
\hfill
\begin{minipage}[t]{0.49\textwidth}
    \vspace{0pt}
    \centering
    \captionof{table}{Success rates for multi-stage manipulation tasks.}
    \label{tab:multi_stage_success_rate}
    \renewcommand{\arraystretch}{1.15}
    \resizebox{\linewidth}{!}{
    \begin{tabular}{llcccccc}
    \toprule
    \multirow{2}{*}{Task}
    & \multirow{2}{*}{Method}
    & \multicolumn{5}{c}{Subtask}
    & \multirow{2}{*}{SR (95\% CI)} \\
    \cmidrule(lr){3-7}
    & & 1 & 2 & 3 & 4 & 5 & \\
    \midrule

    \multirow{5}{*}{T2}
    & $\pi_{0.5}$
    & 11/20 & 11/20 & 8/20 & 4/20 & 4/20
    & 20\% (8.1--41.6) \\
    \cmidrule(l){2-8}

    & State Force + Text Force Goal
    & 17/20 & 16/20 & 11/20 & 4/20 & 4/20
    & 20\% (8.1--41.6) \\
    \cmidrule(l){2-8}

    & Visual Force + Text Force Goal
    & 16/20 & 16/20 & 15/20 & 14/20 & 13/20
    & 65\% (43.3--81.9) \\
    \cmidrule(l){2-8}

    & \textbf{VisForce}
    & \textbf{16/20} & \textbf{16/20} & \textbf{16/20}
    & \textbf{15/20} & \textbf{14/20}
    & \textbf{70\% (48.1--85.5)} \\

    \toprule

    \multirow{5}{*}{T3}
    & $\pi_{0.5}$
    & 14/20 & 4/20 & 4/20 & 1/20 & 1/20
    & 5\% (0.9--23.6) \\
    \cmidrule(l){2-8}

    & State Force + Text Force Goal
    & 14/20 & 5/20 & 5/20 & 1/20 & 1/20
    & 5\% (0.9--23.6) \\
    \cmidrule(l){2-8}

    & Visual Force + Text Force Goal
    & 17/20 & 6/20 & 4/20 & 4/20 & 4/20
    & 20\% (8.1--41.6) \\
    \cmidrule(l){2-8}

    & \textbf{VisForce}
    & \textbf{20/20} & \textbf{14/20} & \textbf{14/20}
    & \textbf{11/20} & \textbf{11/20}
    & \textbf{55\% (34.2--74.2)} \\

    \toprule

    \multirow{5}{*}{T4}
    & $\pi_{0.5}$
    & 5/20 & 1/20 & 1/20 & - & -
    & 5\% (0.9--23.6) \\
    \cmidrule(l){2-8}

    & State Force + Text Force Goal
    & 17/20 & 11/20 & 1/20 & - & -
    & 5\% (0.9--23.6) \\
    \cmidrule(l){2-8}

    & Visual Force + Text Force Goal
    & 20/20 & 18/20 & 4/20 & - & -
    & 20\% (8.1--41.6) \\
    \cmidrule(l){2-8}

    & \textbf{VisForce}
    & \textbf{20/20} & \textbf{16/20} & \textbf{8/20}
    & \textbf{-} & \textbf{-}
    & \textbf{40\% (21.9--61.3)} \\

    \bottomrule
    \end{tabular}
    }
\end{minipage}
\end{figure*}

\textit{T3: Tong-Assisted Bread Transfer.}
T3 corresponds to the Tong-Assisted Bread Transfer task shown in Fig.~\ref{fig:experiment_setup}, where the robot maintains a persistent tong grasp while adapting force across manipulation stages. The robot sequentially transfers large and small bread pieces using the desired forces specified in T3 of Table~\ref{tab:finger_desired_force_multi}. This setting evaluates whether the policy can repeatedly adapt to changing force requirement while preserving the same tool contact.

\textit{T4: Slip-Modulated Peg-in-Hole.}
T4 corresponds to the Slip-Modulated Peg-in-Hole task shown in Fig.~\ref{fig:experiment_setup}. Following the desired forces specified for T4 in Table~\ref{tab:finger_desired_force_multi}, the robot uprights a horizontally placed peg through slip and then increases the force for insertion into a 1-mm clearance hole. This task examines whether the policy can transition from low-force slip to high-force stabilization on the same object.

\subsection{Force-Conditioned Grasping}

We analyze whether changes in the desired force condition are reflected in the actual actuator force in the Rigid Object setting. Fig.~\ref{fig:force_grasping}(a) shows the measured aggregate actuator force under low, medium, and high desired force conditions. Aggregate actuator force is computed as the mean of the active-finger actuator forces during the stable grasp interval. VisForce shows a consistent increase in aggregate actuator force as the desired force increases, with clear separation between the three conditions, whereas the comparison methods show relatively limited differentiation across conditions.

Fig.~\ref{fig:force_grasping}(b) shows grasp-and-lift success on force-sensitive objects under the medium desired-force condition. VisForce achieves success rates of 70\% and 80\% on the egg and toothpaste tasks, respectively, compared with 40\% and 20\% for Visual Force + Text Force Goal, and 35\% and 5\% for State Force + Text Force Goal. The consistent force responses in the rigid-object experiment, together with the success on force-sensitive objects, suggest that VisForce not only differentiates force levels, but also provides task-appropriate force conditioning in real manipulation scenarios where excessive or insufficient grasp force can lead to failure.

\subsection{Multi-Stage Force-Aware Manipulation Results}

Fig.~\ref{fig:subtask_progress} illustrates the stage-wise progression of T2--T4, while Table~\ref{tab:multi_stage_success_rate} compares their stage-wise and final success rates. Values in parentheses indicate 95\% Wilson confidence intervals for the final task success rates.

\textbf{T2: Cup Insertion and Bottle Pouring.}
For T2, Table~\ref{tab:multi_stage_success_rate} reports final success rates of 20\% for both $\pi_{0.5}$ and State Force + Text Force Goal, 65\% for Visual Force + Text Force Goal, and 70\% for VisForce. The stage-wise results further show that Visual Force + Text Force Goal and VisForce maintain relatively high cumulative success through the pouring and placement stages following bottle grasping.

State Force + Text Force Goal and Visual Force + Text Force Goal use the same setup except for the representation of the current force. Under this matched $\pi_{0.5}$-based setup, fingertip-aligned visual grounding of actuator force resulted in higher task success than directly appending the same force values to the proprioceptive state. Fig.~\ref{fig:subtask_progress} shows that, during the bottle power grasp in T2, most fingertips are occluded in the RGB observation. In this setting, the explicit correspondence between visual force cues and fingertip locations may provide additional spatial information about the grasp state despite limited fingertip visibility.

\textbf{T3: Tong-Assisted Bread Transfer.}
Table~\ref{tab:multi_stage_success_rate} reports a final success rate of 55\% for VisForce on T3, compared with 20\% for Visual Force + Text Force Goal. As illustrated by the T3 progression in Fig.~\ref{fig:subtask_progress}, the performance gap becomes more pronounced after bread manipulation begins. While the cumulative success of Visual Force + Text Force Goal drops rapidly after the large-bread grasp, VisForce maintains a higher cumulative success rate throughout the subsequent stages.

The two methods use the same visual grounding of the current force in the wrist image and the same cross-attention mechanism, differing only in how the desired force is represented. The larger performance gap after bread manipulation begins suggests that the difference between the two representations becomes more apparent when the desired force changes repeatedly across manipulation stages rather than maintained as a single force condition. These results indicate that, for T3, where stage-dependent force adjustment is repeatedly required while maintaining a persistent tool grasp, providing the desired force as a goal image with fingertip-aligned visual grounding is more effective than providing it as text.

\textbf{T4: Slip-Modulated Peg-in-Hole.}
For T4 in Table~\ref{tab:multi_stage_success_rate}, $\pi_{0.5}$ and State Force + Text Force Goal achieve final success rates of 5\% each, while Visual Force + Text Force Goal and VisForce achieve 20\% and 40\%, respectively. In the T4 sequence in Fig.~\ref{fig:subtask_progress}, the methods using visual grounding of actuator force maintain relatively high success through the controlled-slip uprighting stage, while VisForce maintains higher cumulative success through the subsequent grasp stabilization and insertion stages.

Compared with Visual Force + Text Force Goal, VisForce shows a higher observed final success rate in this setting. This result suggests that providing the desired force as a fingertip-aligned visual goal can help the policy transition between force-dependent interactions, such as low-force slip and high-force stabilization, on the same object.

\subsection{Ablation of Current--Goal Visual Fusion}

\begin{figure}[t]
    \centering
    \centerline{\includegraphics[width=\columnwidth]{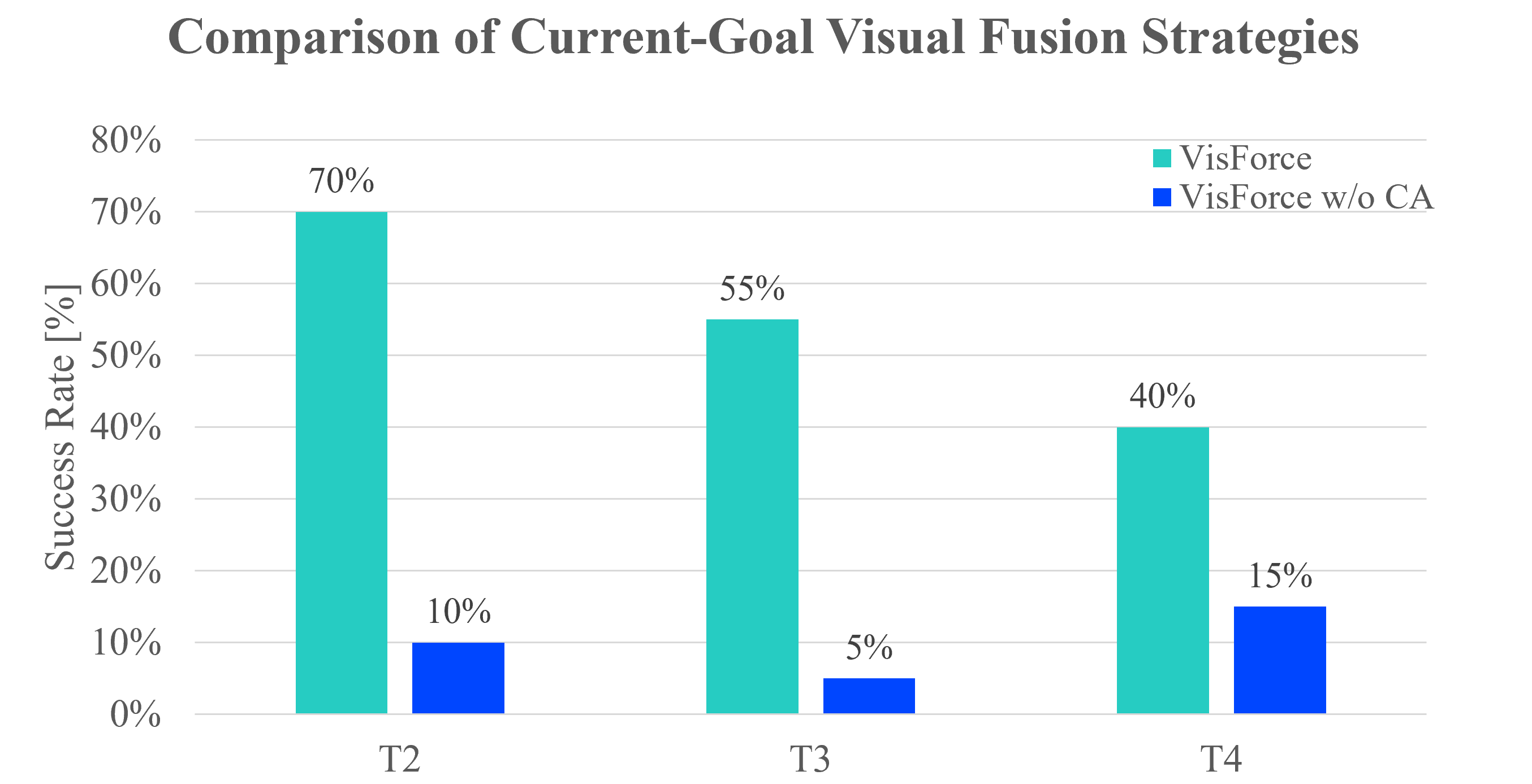}}
    \caption{\textbf{Ablation of Current--Goal Visual Fusion.}
    Both methods receive the same force-grounded wrist image and force-grounded goal image, while differing only in how the current and goal visual tokens are fused. Bars report the final task success rate over 20 trials for each manipulation task.}
    \label{fig:comparison_of_fusion}
\end{figure}

To analyze the effect of the fusion strategy between the current and goal representations, we compare VisForce w/o CA with VisForce. As shown in Fig.~\ref{fig:comparison_of_fusion}, VisForce w/o CA achieves final success rates of 10\%, 5\%, and 15\% on T2, T3, and T4, respectively, whereas VisForce achieves 70\%, 55\%, and 40\%. The two methods use the same force-grounded wrist and goal images and differ only in the fusion strategy.

With simple concatenation, the correspondence between current and goal tokens must be learned implicitly by subsequent VLM layers. In contrast, the proposed cross-attention allows current wrist tokens to attend directly to task-relevant foreground goal tokens, explicitly conditioning the current observation on the goal representation. These results indicate that the proposed goal-conditioned cross-attention is more effective for current--goal fusion than simple visual-token concatenation.

\section{CONCLUSIONS AND DISCUSSION}

We presented VisForce, a force-grounded visual conditioning method for dexterous hand manipulation. VisForce represents the current actuator force and task-specific desired force as image-space cues grounded at the corresponding fingertip locations, and generates actions by conditioning the force-grounded wrist image on the force-grounded goal image through goal-conditioned cross-attention.

In real-robot experiments, VisForce showed distinguishable grip-force responses across different desired-force conditions in the T1 rigid-object setting, and achieved higher task success than the comparison methods in force-sensitive grasping and multi-stage manipulation tasks. In particular, the results support the effectiveness of the proposed visual force grounding and current--goal conditioning under fingertip occlusion and force adjustment while maintaining persistent tool contact.

\addtolength{\textheight}{-3cm}   
\bibliographystyle{IEEEtran}
\bibliography{references}

@inproceedings{octo,
    title={{Octo}: An Open-Source Generalist Robot Policy},
    author = {{Octo Model Team} and Dibya Ghosh and Homer Walke and Karl Pertsch and Kevin Black and Oier Mees and Sudeep Dasari and Joey Hejna and Charles Xu and Jianlan Luo and Tobias Kreiman and {You Liang} Tan and Lawrence Yunliang Chen and Pannag Sanketi and Quan Vuong and Ted Xiao and Dorsa Sadigh and Chelsea Finn and Sergey Levine},
    booktitle = {Proceedings of Robotics: Science and Systems},
    address  = {Delft, Netherlands},
    year = {2024},
}

@inproceedings{rt-2,
  title={{Rt-2}: Vision-language-action models transfer web knowledge to robotic control},
  author={Zitkovich, Brianna and Yu, Tianhe and Xu, Sichun and Xu, Peng and Xiao, Ted and Xia, Fei and Wu, Jialin and Wohlhart, Paul and Welker, Stefan and Wahid, Ayzaan and others},
  booktitle={Conference on Robot Learning},
  pages={2165--2183},
  year={2023},
  organization={PMLR}
}

@inproceedings{dexgraspvla,
  title={{Dexgraspvla}: A vision-language-action framework towards general dexterous grasping},
  author={Zhong, Yifan and Huang, Xuchuan and Li, Ruochong and Zhang, Ceyao and Chen, Zhang and Guan, Tianrui and Zeng, Fanlian and Lui, Ka Nam and Ye, Yuyao and Liang, Yitao and others},
  booktitle={Proceedings of the AAAI Conference on Artificial Intelligence},
  volume={40},
  number={22},
  pages={18836--18844},
  year={2026}
}

@inproceedings{openvla,
  title={{OpenVLA}: An Open-Source Vision-Language-Action Model},
  author={Kim, Moo Jin and Pertsch, Karl and Karamcheti, Siddharth and Xiao, Ted and Balakrishna, Ashwin and Nair, Suraj and Rafailov, Rafael and Foster, Ethan P and Sanketi, Pannag R and Vuong, Quan and others},
  booktitle={Conference on Robot Learning},
  pages={2679--2713},
  year={2025},
  organization={PMLR}
}

@inproceedings{pi_0,
  title     = {$\pi_0$: A Vision-Language-Action Flow Model for General Robot Control},
  author    = {Black, Kevin and Brown, Noah and Driess, Danny and Esmail, Adnan
               and Equi, Michael Robert and Finn, Chelsea and Fusai, Niccolo
               and Groom, Lachy and Hausman, Karol and Ichter, Brian
               and Jakubczak, Szymon and Jones, Tim and Ke, Liyiming
               and Levine, Sergey and Li-Bell, Adrian and Mothukuri, Mohith
               and Nair, Suraj and Pertsch, Karl and Shi, Lucy Xiaoyang
               and Smith, Laura and Tanner, James and Vuong, Quan
               and Walling, Anna and Wang, Haohuan and Zhilinsky, Ury},
  booktitle = {Proceedings of Robotics: Science and Systems},
  year      = {2025},
  doi       = {10.15607/RSS.2025.XXI.010}
}

@inproceedings{pi_05,
  title     = {$\pi_{0.5}$: a Vision-Language-Action Model with Open-World Generalization},
  author    = {Black, Kevin and Brown, Noah and Darpinian, James
               and Dhabalia, Karan and Driess, Danny and Esmail, Adnan
               and Equi, Michael Robert and Finn, Chelsea and Fusai, Niccolo
               and Galliker, Manuel Y. and Ghosh, Dibya and Groom, Lachy
               and Hausman, Karol and Ichter, Brian and Jakubczak, Szymon
               and Jones, Tim and Ke, Liyiming and LeBlanc, Devin
               and Levine, Sergey and Li-Bell, Adrian and Mothukuri, Mohith
               and Nair, Suraj and Pertsch, Karl and Ren, Allen Z.
               and Shi, Lucy Xiaoyang and Smith, Laura
               and Springenberg, Jost Tobias and Stachowicz, Kyle
               and Tanner, James and Vuong, Quan and Walke, Homer
               and Walling, Anna and Wang, Haohuan and Yu, Lili
               and Zhilinsky, Ury},
  booktitle = {Proceedings of The 9th Conference on Robot Learning},
  pages     = {17--40},
  year      = {2025},
  volume    = {305},
  series    = {Proceedings of Machine Learning Research},
  publisher = {PMLR}
}

@inproceedings{goalrep,
  title={Goal representations for instruction following: A semi-supervised language interface to control},
  author={Myers, Vivek and He, Andre Wang and Fang, Kuan and Walke, Homer Rich and Hansen-Estruch, Philippe and Cheng, Ching-An and Jalobeanu, Mihai and Kolobov, Andrey and Dragan, Anca and Levine, Sergey},
  booktitle={Conference on Robot Learning},
  pages={3894--3908},
  year={2023},
  organization={PMLR}
}

@inproceedings{foreact,
  title={{Foreact}: Steering your vla with efficient visual foresight planning},
  author={Zhang, Zhuoyang and Yang, Shang and Hu, Qinghao and Huang, Luke J and Hou, James and Sun, Yufei and Lu, Yao and Han, Song},
  booktitle={Proceedings of the IEEE/CVF Conference on Computer Vision and Pattern Recognition},
  pages={37195--37205},
  year={2026}
}

@inproceedings{cot,
  title={{CoT-VLA}: Visual Chain-of-Thought Reasoning for Vision-Language-Action Models},
  author={Zhao, Qingqing and Lu, Yao and Kim, Moo Jin and Fu, Zipeng and Zhang, Zhuoyang and Wu, Yecheng and Li, Zhaoshuo and Ma, Qianli and Han, Song and Finn, Chelsea and others},
  booktitle={Proceedings of the IEEE/CVF Conference on Computer Vision and Pattern Recognition},
  pages={1702--1713},
  year={2025},
  organization={IEEE}
}

@inproceedings{susie,
  title={Zero-shot robotic manipulation with pre-trained image-editing diffusion models},
  author={Black, Kevin and Nakamoto, Mitsuhiko and Atreya, Pranav and Walke, Homer and Finn, Chelsea and Kumar, Aviral and Levine, Sergey},
  booktitle={International Conference on Learning Representations},
  volume={2024},
  pages={33431--33452},
  year={2024}
}

@inproceedings{forediffusion,
  title={{ForeDiffusion}: Foresight-Conditioned Diffusion Policy via Future View Construction for Robot Manipulation},
  author={Xie, Weize and Ding, Yi and He, Ying and Wang, Leilei and Bai, Binwen and Zhao, Zheyi and Wang, Chenyang and Yu, F Richard},
  booktitle={Proceedings of the AAAI Conference on Artificial Intelligence},
  volume={40},
  number={22},
  pages={18665--18673},
  year={2026}
}

@article{goal_vla,
  title={{Goal-VLA}: Image-Generative VLMs as Object-Centric World Models Empowering Zero-shot Robot Manipulation},
  author={Chen, Haonan and Guo, Jingxiang and Wang, Bangjun and Zhang, Tianrui and Huang, Xuchuan and Zheng, Boren and Hou, Yiwen and Tie, Chenrui and Deng, Jiajun and Shao, Lin},
  journal={arXiv preprint arXiv:2506.23919},
  year={2025}
}

@article{forcevla,
  title={{ForceVLA}: Enhancing vla models with a force-aware moe for contact-rich manipulation},
  author={Yu, Jiawen and Liu, Hairuo and Yu, Qiaojun and Ren, Jieji and Hao, Ce and Ding, Haitong and Huang, Guangyu and Huang, Guofan and Song, Yan and Cai, Panpan and others},
  journal={Advances in Neural Information Processing Systems},
  volume={38},
  pages={93409--93439},
  year={2025}
}

@article{taf_vla,
  title={Tactile-Force Alignment in Vision-Language-Action Models for Force-aware Manipulation},
  author={Huang, Yuzhe and Lin, Pei and Li, Wanlin and Li, Daohan and Li, Jiajun and Jiang, Jiaming and Xiao, Chenxi and Jiao, Ziyuan},
  journal={arXiv preprint arXiv:2601.20321},
  year={2026}
}

@article{fd_vla,
  title={{FD-VLA}: Force-Distilled Vision-Language-Action Model for Contact-Rich Manipulation},
  author={Zhao, Ruiteng and Wang, Wenshuo and Ma, Yicheng and Li, Xiaocong and Tay, Francis EH and Ang Jr, Marcelo H and Zhu, Haiyue},
  journal={arXiv preprint arXiv:2602.02142},
  year={2026}
}

@inproceedings{forcesight,
  author={Collins, Jeremy A. and Houff, Cody and Tan, You Liang and Kemp, Charles C.},
  booktitle={2024 IEEE International Conference on Robotics and Automation (ICRA)}, 
  title={ForceSight: Text-Guided Mobile Manipulation with Visual-Force Goals}, 
  year={2024},
  volume={},
  number={},
  pages={10874-10880},
  doi={10.1109/ICRA57147.2024.10611210}
}

@article{vla_touch,
  title={{VLA-Touch}: Enhancing Vision-Language-Action Model with Dual-Level Tactile Feedback},
  author={Bi, Jianxin and Ma, Kevin Yuchen and Hao, Ce and Shou, Mike Zheng and Soh, Harold},
  journal={IEEE Robotics and Automation Letters},
  volume={11},
  number={7},
  pages={8487--8494},
  year={2026},
  doi={10.1109/LRA.2026.3692345}
}

@article{tactile_vla,
  title={{Tactile-VLA}: unlocking vision-language-action model's physical knowledge for tactile generalization},
  author={Huang, Jialei and Wang, Shuo and Lin, Fanqi and Hu, Yihang and Wen, Chuan and Gao, Yang},
  journal={arXiv preprint arXiv:2507.09160},
  year={2025}
}

@article{johansson2009coding,
  title={Coding and use of tactile signals from the fingertips in object manipulation tasks},
  author={Johansson, Roland S and Flanagan, J Randall},
  journal={Nature Reviews Neuroscience},
  volume={10},
  number={5},
  pages={345--359},
  year={2009},
  publisher={Nature Publishing Group UK London}
}

@article{schettino2003effects,
  title={Effects of object shape and visual feedback on hand configuration during grasping},
  author={Schettino, Luis F and Adamovich, Sergei V and Poizner, Howard},
  journal={Experimental Brain Research},
  volume={151},
  number={2},
  pages={158--166},
  year={2003},
  publisher={Springer}
}

@inproceedings{todorov2012mujoco,
  title={Mujoco: A physics engine for model-based control},
  author={Todorov, Emanuel and Erez, Tom and Tassa, Yuval},
  booktitle={2012 IEEE/RSJ international conference on intelligent robots and systems},
  pages={5026--5033},
  year={2012},
  organization={IEEE}
}

@article{storn1997differential,
  title={Differential evolution--a simple and efficient heuristic for global optimization over continuous spaces},
  author={Storn, Rainer and Price, Kenneth},
  journal={Journal of global optimization},
  volume={11},
  number={4},
  pages={341--359},
  year={1997},
  publisher={Springer}
}

@article{powell1964efficient,
  title={An efficient method for finding the minimum of a function of several variables without calculating derivatives},
  author={Powell, Michael JD},
  journal={The computer journal},
  volume={7},
  number={2},
  pages={155--162},
  year={1964},
  publisher={Oxford University Press}
}

@inproceedings{sam2,
  title={{Sam 2}: Segment anything in images and videos},
  author={Ravi, Nikhila and Gabeur, Valentin and Hu, Yuan-Ting and Hu, Ronghang and Ryali, Chaitanya and Ma, Tengyu and Khedr, Haitham and R{\"a}dle, Roman and Rolland, Chloe and Gustafson, Laura and others},
  booktitle={International Conference on Learning Representations},
  volume={2025},
  pages={28085--28128},
  year={2025}
}

@inproceedings{groundingdino,
  title={Grounding dino: Marrying dino with grounded pre-training for open-set object detection},
  author={Liu, Shilong and Zeng, Zhaoyang and Ren, Tianhe and Li, Feng and Zhang, Hao and Yang, Jie and Jiang, Qing and Li, Chunyuan and Yang, Jianwei and Su, Hang and others},
  booktitle={European conference on computer vision},
  pages={38--55},
  year={2024},
  organization={Springer}
}

@article{dexforce,
  title={{DexForce}: Extracting Force-Informed Actions From Kinesthetic Demonstrations for Dexterous Manipulation},
  author={Chen, Claire and Yu, Zhongchun and Choi, Hojung and Cutkosky, Mark and Bohg, Jeannette},
  journal={IEEE Robotics and Automation Letters},
  volume={10},
  number={6},
  pages={6416--6423},
  year={2025}
}

@inproceedings{kinedex,
  title={{KineDex}: Learning Tactile-Informed Visuomotor Policies via Kinesthetic Teaching for Dexterous Manipulation},
  author={Zhang, Di and Yuan, Chengbo and Wen, Chuan and Zhang, Hai and Zhao, Junqiao and Gao, Yang},
  booktitle={Proceedings of The 9th Conference on Robot Learning},
  pages={4123--4138},
  year={2025},
  volume={305},
  series={Proceedings of Machine Learning Research},
  publisher={PMLR}
}

@article{foar,
  title={{FoAR}: Force-Aware Reactive Policy for Contact-Rich Robotic Manipulation},
  author={He, Zihao and Fang, Hongjie and Chen, Jingjing and Fang, Hao-Shu and Lu, Cewu},
  journal={IEEE Robotics and Automation Letters},
  volume={10},
  number={6},
  pages={5625--5632},
  year={2025}
}

@inproceedings{hu2021lora,
  title     = {{LoRA}: Low-Rank Adaptation of Large Language Models},
  author    = {Hu, Edward J. and Shen, Yelong and Wallis, Phillip and Allen-Zhu, Zeyuan
               and Li, Yuanzhi and Wang, Shean and Wang, Lu and Chen, Weizhu},
  booktitle = {International Conference on Learning Representations},
  year      = {2022}
}

@inproceedings{rahman2016iou,
  title={Optimizing intersection-over-union in deep neural networks for image segmentation},
  author={Rahman, Md Atiqur and Wang, Yang},
  booktitle={International symposium on visual computing},
  pages={234--244},
  year={2016},
  organization={Springer}
}

@article{borgefors1988chamfer,
  title={Hierarchical chamfer matching: A parametric edge matching algorithm},
  author={Borgefors, Gunilla},
  journal={IEEE Transactions on pattern analysis and machine intelligence},
  volume={10},
  number={6},
  pages={849--865},
  year={1988},
  publisher={IEEE}
}

@article{fan2026peafowl,
  title={{Peafowl}: Perception-enhanced multi-view vision-language-action for bimanual manipulation},
  author={Fan, Qingyu and Li, Zhaoxiang and Hu, Jinrui and Lu, Yi and Chen, Wang and Shen, Qiu and Long, Xiao-xiao and Cai, Yinghao and Lu, Tao and Wang, Shuo and others},
  journal={IEEE Robotics and Automation Letters},
  year={2026},
  publisher={IEEE}
}

@article{visual_force_tactile,
  title={Visual-force-tactile fusion for gentle intricate insertion tasks},
  author={Jin, Piaopiao and Huang, Bidan and Lee, Wang Wei and Li, Tiefeng and Yang, Wei},
  journal={IEEE Robotics and Automation Letters},
  volume={9},
  number={5},
  pages={4830--4837},
  year={2024},
  publisher={IEEE}
}

@article{implicitrdp,
  title={{ImplicitRDP}: An end-to-end visual-force diffusion policy with structural slow-fast learning},
  author={Chen, Wendi and Xue, Han and Wang, Yi and Zhou, Fangyuan and Lv, Jun and Jin, Yang and Tang, Shirun and Wen, Chuan and Lu, Cewu},
  journal={IEEE Robotics and Automation Letters},
  year={2026},
  publisher={IEEE}
}

@article{pocodp3,
  title={{PoCoDP3}: Pose-and Contact-Aware Visual-Tactile Policy for Contact-Rich 3D Manipulation},
  author={Yue, Zhaokun and Tong, Ling and Qian, Kun},
  journal={IEEE Robotics and Automation Letters},
  volume={11},
  number={3},
  pages={2434--2441},
  year={2026},
  publisher={IEEE}
}

@article{grmg,
  title={{Gr-mg}: Leveraging partially-annotated data via multi-modal goal-conditioned policy},
  author={Li, Peiyan and Wu, Hongtao and Huang, Yan and Cheang, Chilam and Wang, Liang and Kong, Tao},
  journal={IEEE Robotics and Automation Letters},
  volume={10},
  number={2},
  pages={1912--1919},
  year={2025},
  publisher={IEEE}
}

@article{dextouch,
  title={{DexTouch}: Learning to Seek and Manipulate Objects With Tactile Dexterity},
  author={Lee, Kang-Won and Qin, Yuzhe and Wang, Xiaolong and Lim, Soo-Chul},
  journal={IEEE Robotics and Automation Letters},
  volume={9},
  number={12},
  pages={10772--10779},
  year={2024},
  doi={10.1109/LRA.2024.3478571}
}
\end{document}